\documentclass{article}

    \PassOptionsToPackage{numbers, compress}{natbib}

\usepackage[preprint]{neurips_2020}

\usepackage[utf8]{inputenc} 
\usepackage[T1]{fontenc}    
\usepackage{hyperref}       
\usepackage{url}            
\usepackage{xcolor}
\usepackage{booktabs}       
\usepackage{amsfonts}       
\usepackage{nicefrac}       
\usepackage{microtype}      
\usepackage{ulem}
\usepackage{algorithm}
\usepackage{algpseudocode}
\usepackage{float}
\usepackage{caption}
\usepackage{subcaption}
 \usepackage{graphics}
\usepackage{amssymb,graphicx}
\usepackage{amsmath}
\usepackage{soul}

\title{Improved knowledge distillation by utilizing backward pass knowledge in neural networks}

\author{%
  Aref Jafari\thanks{This work was done while doing internship at Huawei Noah’s Ark Lab}\\
  
  University of Waterloo\\
  \texttt{aref.jafari@uwaterloo.ca} \\
  \And
  Mehdi Rezagholizadeh \\
  Huawei Noah’s Ark Lab \\
   \texttt{mehdi.rezagholizadeh@huawei.com} \\
   \AND
  Ali Ghodsi \\
  University of Waterloo \\
  \texttt{ali.ghodsi@uwaterloo.ca} \\
}

\begin{document}

\maketitle

\begin{abstract}
Knowledge distillation (KD) is one of the prominent techniques for model compression. In this method, the knowledge of a large network (teacher) is distilled into a model (student) with usually significantly fewer parameters. KD tries to better-match the output of the student model to that of the teacher model based on the knowledge extracts from the forward pass of the teacher network. Although conventional KD is effective for matching the two networks over the given data points, there is no guarantee that these models would match in other areas for which we do not have enough training samples. In this work, we address that problem by generating new auxiliary training samples based on extracting knowledge from the backward pass of the teacher in the areas where the student diverges greatly from the teacher. We compute the difference between the teacher and the student and generate new data samples that maximize the divergence. This is done by perturbing data samples in the direction of the gradient of the difference between the student and the teacher. Augmenting the training set by adding this auxiliary improves the performance of KD significantly and leads to a closer match between the student and the teacher. Using this approach, when data samples come from a discrete domain, such as applications of natural language processing (NLP) and language understanding, is not trivial. However, we show how this technique can be used successfully in such applications. We evaluated the performance of our method on various tasks in computer vision and NLP domains and got promising results.

\end{abstract}

\section{Introduction}

During the last few years, we faced the emerge of a huge number of cumbersome state-of-the-art deep neural network models in different fields of machine learning, including computer vision~\cite{wong2019yolo,howard2017mobilenets}, natural language processing~\cite{prato2019fully,jiao2019tinybert,lan2019albert,brown2020language} and speech processing~\cite{bie2019fully,he2019streaming}.
We need powerful servers to be able to deploy such large models. Running such large models on edge devices would be infeasible due to the limited memory and computational power of edge devices~\cite{sun2020mobilebert}. On the other hand, considering users' privacy concerns, network reliability issues, and network delays increase the demand for offline machine learning solutions on edge devices. The field of neural model compression focuses on providing compression solutions such as quantization~\cite{jacob2018quantization}, pruning~\cite{wang2019pruning}, tensor decomposition~\cite{tjandra2018tensor} and knowledge distillation (KD)~\cite{hinton2015distilling} for large neural networks.   

\begin{figure}[tb]
 \centering
  \subfloat[]{\label{figur:2}\includegraphics[width=0.4\textwidth]{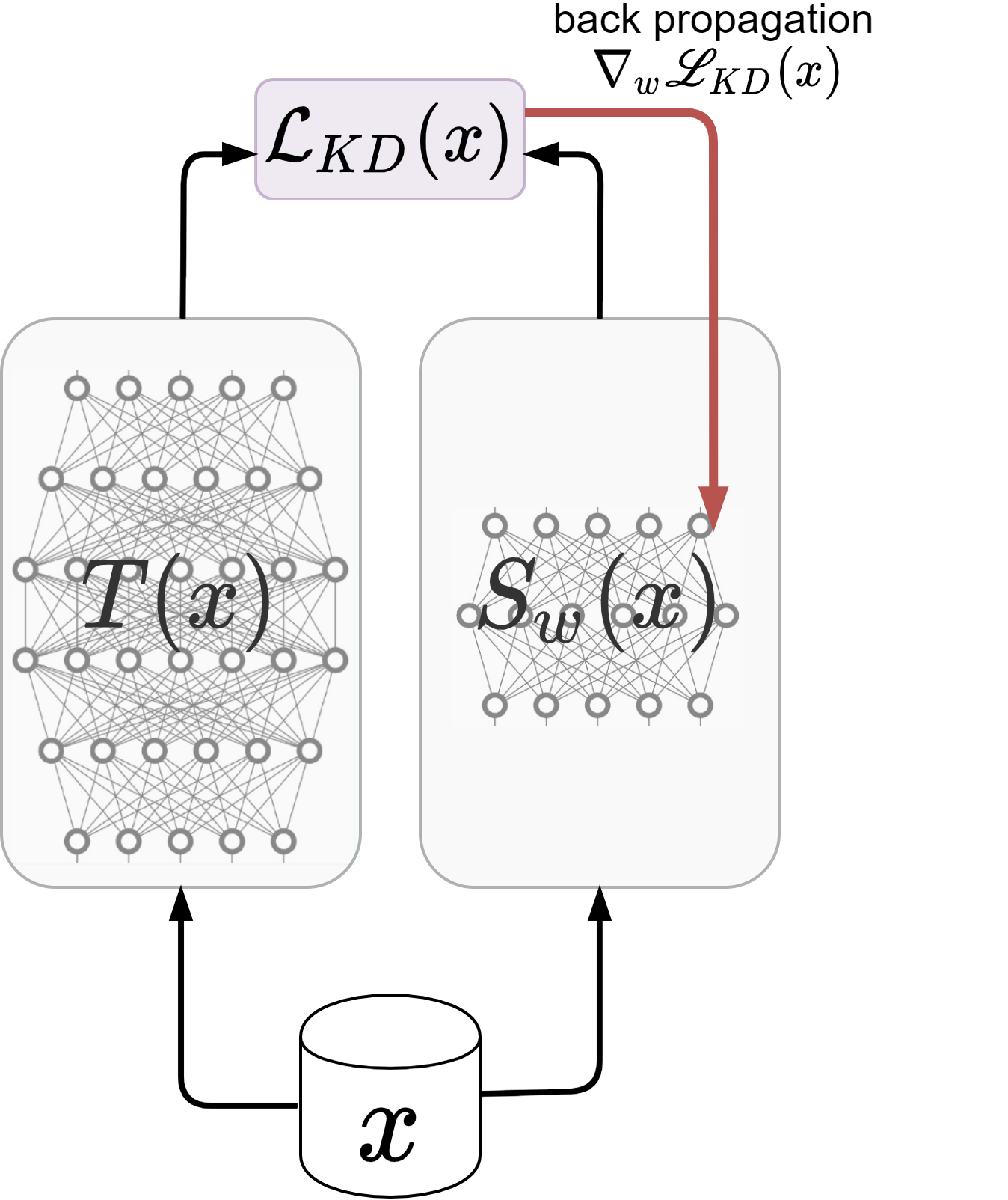}} 
  \subfloat[]{\label{figur:1}\includegraphics[width=0.47\textwidth]{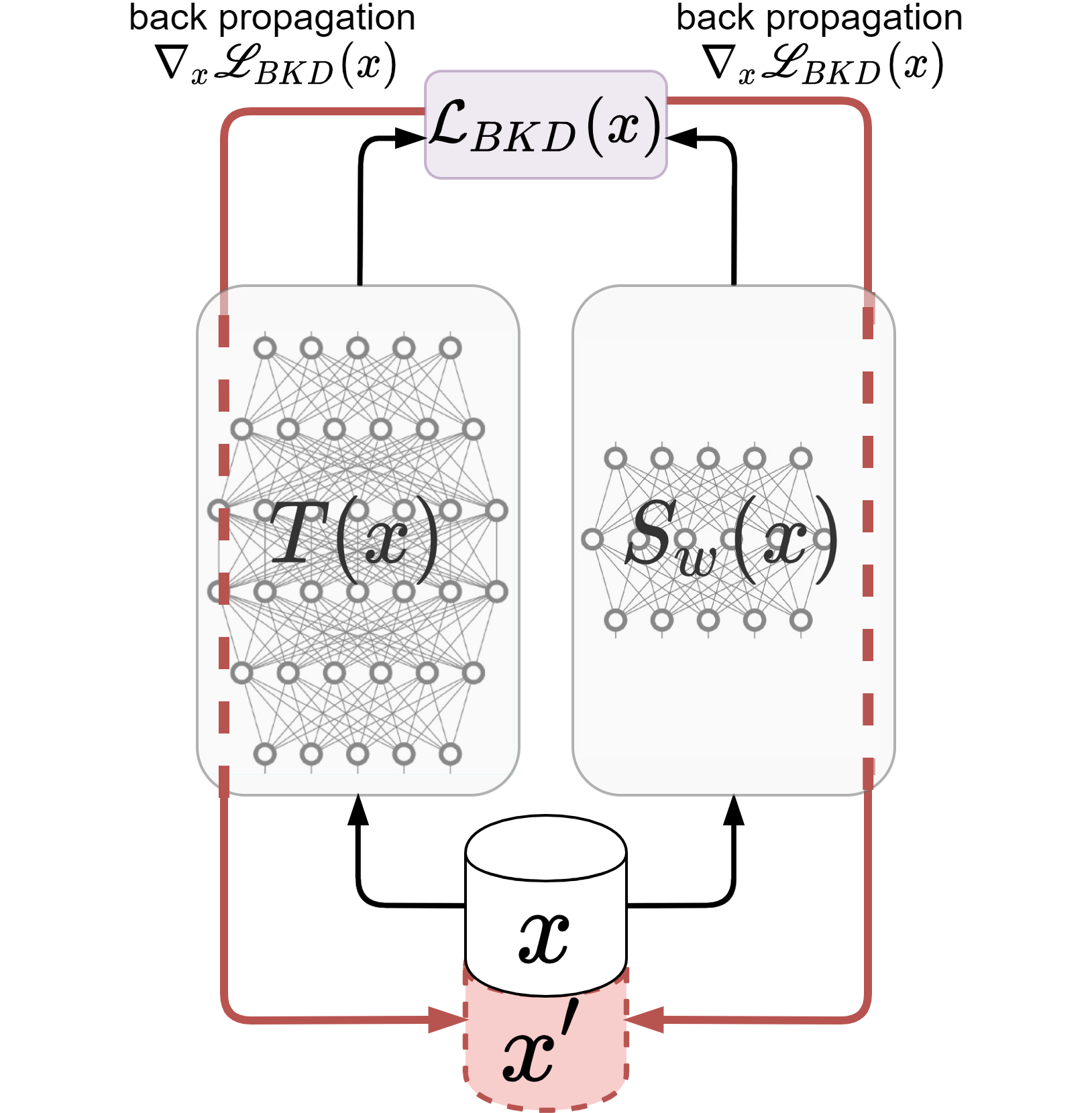}}
  \label{figur}\caption{(a) Minimization Step: Using the teacher model knowledge for training the student in KD (utilizing forward knowledge) (b) Maximization Step: Augmenting the input dataset $x$ with auxiliary data samples $x'$ which is generated by the back propagation of gradient through both networks (utilizing backward knowledge)}
\end{figure}

Knowledge distillation (KD) is one of the most prominent compression techniques in the literature. As its name implies, KD tries to transfer the learned knowledge from a large teacher network to a small student. 
The idea of {KD} {was proposed by Rich Caruana et al. ~\cite{bucilua2006model} for the first time and later this idea generalized by}  Hinton et al.  2015~\cite{hinton2015distilling} for deep neural nets. The original KD method concerns transferring knowledge from a teacher to a student network only by matching their forward pass outputs. Later on, several works in the literature suggested other sources of knowledge in the teacher network besides the logit outputs of the last layer. This includes using intermediate layer feature maps~\cite{sun2019patient,sunmobilebert, jiao2019tinybert}, gradients of the network outputs w.r.t the inputs~\cite{2017arXiv170604859C, 2018arXiv180300443S}), and matching decision boundaries for classification tasks~~\cite{heo2019knowledge}. using this additional information might be useful to get the student network performance closer to that of the teacher. 

In this work, we focus on identifying regions of the input space of the teacher and student networks in which the two functions diverge the most from each other. Moreover, we highlight the importance of incorporating backward knowledge of the teacher and student networks in the knowledge distillation process. Our proposed iterative backward KD approach is comprised of: first, a maximization step in which a new set of auxiliary training samples is generated by pushing training samples towards maximum divergence regions of the two functions; second, a minimization step in which the student network is trained using the regular KD approach over its training data together with the generated auxiliary samples from the first step. We show the success of our backward KD technique in improving KD on both classification and regression tasks over the image and textual data and also in the few-sample KD scenario.  
We summarize the main contributions of this paper in the following: 
\begin{itemize}
\item Our technique extracts knowledge from both the forward and backward passes of the teacher and student networks in order to identify the maximum divergence regions between the two functions and generate auxiliary data samples around those regions. 
\item We provide a solution on how to address the non-differentiability of discrete tokens in NLP applications. 
\item Our approach is generic and is applicable to any improved KD approach.  
\item The results of our experiments, show 4\% improvement on MNIST with a student network that is 160 times smaller, 1\% improvement on the CIFAR-10 dataset with a student that is 9 times smaller, and an average 1.5\% improvement on the GLUE benchmark with a distilroBERTa-base student.
\end{itemize}



\section{Related Works}
\label{related}

\subsection{Knowledge Distillation}
\label{KD}


In the original KD, the process of transferring knowledge from a teacher to a student model accomplishes by minimizing a loss function between the logits of student and teacher networks. This loss function has been used in addition to the regular training loss function for the student network. In other words, we have an additional loss term in the KD loss function between the softmax outputs of teacher and student networks which is softened by a temperature term. 

\begin{equation}
   \mathcal{L}_{KD} = \alpha \mathcal{L}\bigg(\text{Softmax}\big(S(x)\big), y\bigg) + (1-\alpha) \mathcal{L}\Bigg(\text{Softmax}\bigg(\frac{S(x)}{\tau}\bigg), \text{Softmax}\bigg(\frac{T(x)}{\tau}\bigg)\Bigg) 
   \label{kd_loss}
\end{equation}
where $S(x)$ and $T(x)$ are student and teacher networks respectively. $\tau$ is the temperature parameter and $\alpha$ is a coefficient between $[0,1]$. 
This loss function is a linear combination of two loss functions. The first loss function minimizes the difference between the output of the student model and the given true label. 
The second loss function minimizes the difference between the outputs of the student model and the teacher model.
Therefore the whole loss function minimizes the distance between the student and both underlying and teacher functions. Since the teacher network is assumed to be a good approximation of the underlying function, it should be close enough to the underlying function of data samples. 
Fig.~\ref{fig2}-(a) shows a simple example with three data points, an underlying function, a trained teacher and a potential student function that satisfies the KD loss function in eq.~\ref{kd_loss}.  
However, the problem is, even though the student satisfies the KD objective function and intersects the teacher function close to the training data samples, there is no guarantee that it would fit the teacher network in other regions of the input space as well. In this work, we try to address this problem by deploying the backward gradient information w.r.t the input (we refer to as backward knowledge) in the two networks. 

\subsection{Sobolev Training for KD}

\begin{figure}[hbt]
\centering
    \includegraphics[width=0.6\textwidth]{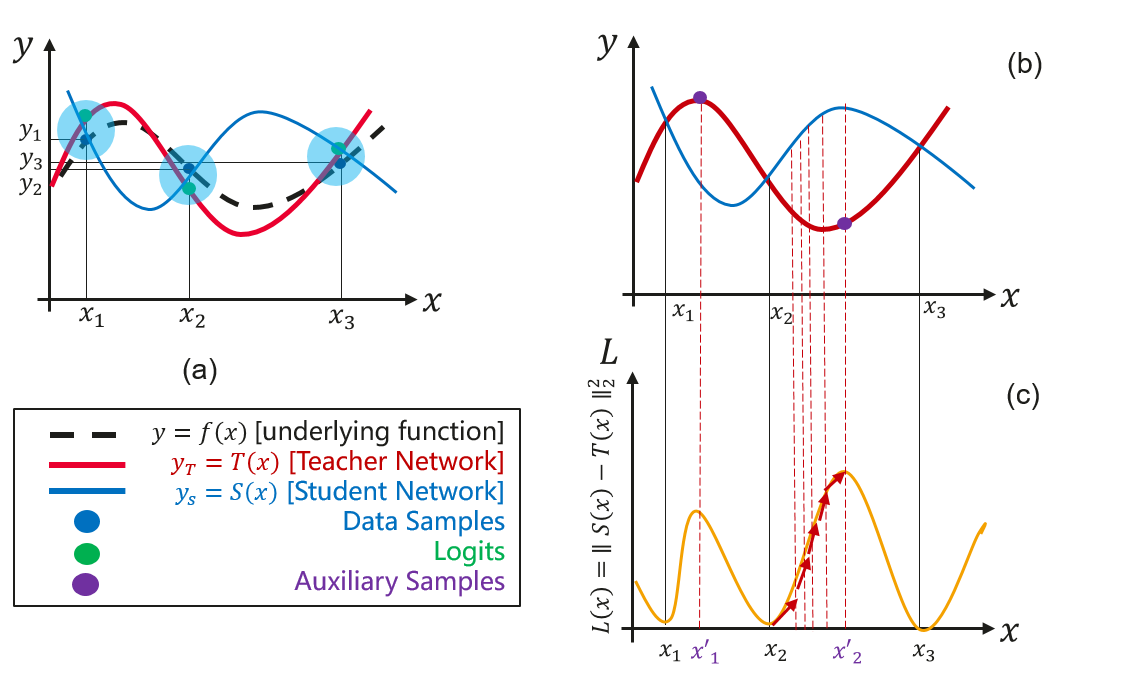}
    \caption{Visualizing the data insufficiency issue for the original KD algorithm. (a) behaviour of the teacher and the student function when training with KD loss. (b) divergence areas between the teacher and the student networks. (c) behaviour of $l_2-norm$ loss function between teacher and the student and the way of obtaining auxiliary data samples.}
    \label{fig2}
\end{figure}

As we mentioned in \ref{KD} (see Fig.~\ref{fig2}.), the KD loss cannot guarantee the student and teacher functions to match over the entire input space.   
The reason is training two networks based on the original KD loss function would only match their output values on the training samples and not their gradients. 
There are some work in the literature to address this issue by matching the gradients of the two networks at given training samples during training \cite{2017arXiv170604859C, 2018arXiv180300443S}.
However, since we usually deal with networks with multidimensional inputs and outputs, the gradients of output vectors w.r.t input vectors give rise to large Jacobin matrices. Matching these Jacobian matrices is not computationally efficient and is not practical in real-world problems.

Sobolev training \cite{2017arXiv170604859C} proposes a solution to avoid large Jacobian matrices: instead of directly matching the gradients of the two networks, one can match the projection of the gradients onto a random vector $v$ which is sampled uniformly from the unit sphere. Although this approach can reduce the computational complexity of matching gradients during the training, still computing Jacobian matrices before this projection can be very computationally expensive (especially for NLP applications that deal with large vocabulary sizes).
To tackle this problem in our work, we define a new scalar loss function based on an $l_2$ norm to measure the distance between the teacher and student networks (see Fig.~\ref{fig2}-(c)). Gradients of this scalar loss function is a vector with the same size as the input vector $x$ and can be used as a proxy for the network gradients introduced in \cite{2017arXiv170604859C, 2018arXiv180300443S}.




\section{Methodology: {Improving Knowledge Distillation using Backward Pass Knowledge} }

In this section, we propose our improved KD method based on generating new out of sample points around the areas of the input domain where the student output diverges greatly from the teacher.
This approach identifies the areas of the input space $\mathcal{X}$ around which the two functions have maximum distance. Then we generate out of sample points $X' \subset \mathcal{X}$ from the existing training set $X \subset \mathcal{X}$ over those regions. These new generated samples $X'$ can be labelled by the teacher and then $ X \leftarrow  X \cup X'$ be deployed in the KD's training process to match the student better to the teacher over a broader range in the input space (see Fig.~\ref{fig2}). We show that augmenting the training set by adding this auxiliary set improves the performance of KD significantly and leads to a closer match between the student and teacher. Our improved KD approach follows a procedure similar to the $minimax$ principle \cite{bratko1982error} : first, in the maximization step we generate auxiliary data samples; second, in the minimization step we apply regular KD on the union of existing $X$ and generated auxiliary data $X'$.

To have a better understanding of how this can be cast as an instance of minimax estimator, assume that we are given the data samples $\{x_i, T(x_i))\}_{i=1}^N$. The goal is to estimate $T(x)$ by $S(x)$. 
We may seek an estimator $S(x)$ attaining the $minimax$ principle. In minimax principle, where $\theta$ is an estimand and $\delta$ is an estimator, we evaluate all estimators according to its maximum risk $R(\theta, \delta)$. An estimator $\delta_0$ , then, is said to be $minimax$ if:
\begin{equation}
    \sup_{\theta} R(\theta, \delta_0) = \inf_{\delta \in C}\sup_{\theta \in \Theta} R(\theta, \delta)
\end{equation}
That is we chose the estimator for the situation that the worst divergence between $\theta$ and $\delta$ is smallest.  We follow a similar insight: i.e.  the maximization step computes $X'$,  where there is the worst divergence between the teacher and the student.  The minimization step finds the weights of the student network such that the difference between the student and teacher for this worst scenario is the smallest.  
\begin{equation}
    \min_{w}\max_{x} R(T_x, S_{x,w})
\end{equation}



    

\subsection{Maximization Step: Generating Auxilary Data based on Backward-KD Loss}


In the maximization step of our technique, we define a new loss function (we refer to as the backward KD loss or BKD throughout this paper) to measure the distance between the output of the teacher and the student networks:   
\begin{equation}
\begin{split}
 \mathcal{L_{BKD}} = ||S(x)-T(x) ||_2^2 
\end{split}
\label{loss_bkd}
\end{equation}
Here the main idea is that by taking the gradient of $\mathcal{L_{BKD}}$ loss function in eq.~\ref{loss_bkd} w.r.t the input samples, we can perturb the training samples along the directions of their gradients to increase the loss between two networks. Using this process, we can generate new auxiliary training samples for which the student and the teacher networks are in maximum distance.  
To obtain these auxiliary data samples, we can consider the following optimization problem. 
\begin{equation}
\begin{split}
    x'=\underset{x\in\mathcal{X}}{\text{max }} ||S(x)-T(x) ||_2^2
\end{split}
\label{opt_bkd}
\end{equation}
We can solve this problem using stochastic gradient ascent method. Therefore our perturbation formula for each data sample will be:
\begin{equation}
    x^{i+1}=x^i+\eta \ \nabla_x \ ||S(x)-T(x) ||_2^2
    \label{iter}
\end{equation}
where in this formula $\eta$ is the perturbation rate. This is an iterative algorithm and $i$ is the iteration index. $x^i$ is a training sample at $i^{th}$ iteration. {Each time, }we perturb $x^i$ by adding a portion of the gradient of loss to this sample. For more detail about this algorithm consider algorithm 1 in the Appendix. 

Fig.~\ref{fig2} demonstrates our idea using a simple example more clearly. Fig.~\ref{fig2}-(a) shows a trained teacher and student functions given the training samples {($x_1$,$y_1$), ($x_2$,$y_2$), ($x_3$,$y_3$)}. 
Fig.~\ref{fig2}-(c) constructs the $\mathcal{L}_{BKD}$  between these two networks. $\mathcal{L}_{BKD}$ shows where the two networks diverge in the original space. Bear in mind that $\mathcal{L}_{BKD}$ gives a scalar for each input. Hence, the gradient of $\mathcal{L}_{BKD}$ with respect to input variable $x$ will be a vector with the same size as the variable $x$. Therefore, it does not need to deal with the large dimensionality issue of the Jacobian matrix as described in ~\cite{2017arXiv170604859C}. 
Fig.~\ref{fig2}-(c) also illustrates an example of generating one auxiliary sample from the training sample $x_2$. 
If we apply eq.~\ref{iter} to this sample, after several iterations, we will reach to a new auxiliary data point ($x_2'$). It is evident in Fig.~\ref{fig2}-(a) that, as expected, there is a large divergence between the teacher and student networks in ($x_2'$) point.

\subsection{Minimization Step: Improving KD with Generated Auxiliary Data  }
\label{maximization}
We can apply the maximization step to all given training data to generate their corresponding auxiliary samples. Then by adding the auxiliary samples $X'$ into the training dataset $X\leftarrow X' \cup X$, we can train the student network again based on the original KD algorithm over the updated training set in order to obtain a better output match between the student and teacher networks. Inspired by~\cite{mirzadeh2019improved}, we have used the following KD loss function in our work:  
\begin{equation}
    {\mathcal{L}_{KD} = (1-\lambda)~H\Big(\sigma\big(S(x)\big),y\Big) + \tau^{2}~\lambda~ KL\bigg(\sigma\Big(\frac{S(x)}{\tau}\Big), \sigma\Big(\frac{T(x)}{\tau}\Big)\bigg)}
\end{equation}
where $\sigma(.)$ is the softmax function, $H(.)$ is the cross-entropy loss function, $KL(.)$ is the Kullback Leibler divergence, $\lambda$ is a hyper parameter, $\tau$ is the temperature parameter, and $y$ is the true labels. 

The intuition behind expecting to get a better KD performance using the updated training data is as follows. 
Now given the auxiliary data samples which point toward the regions of the input space where the student and teacher have maximum divergence, these regions of input space are not dark for the original KD algorithm anymore. 
Therefore, it is expected from the KD algorithm to be able to match the student to the teacher network over a larger input space (see Fig.~\ref{synthetic}).  
Moreover, it is worth mentioning that the maximization and minimization steps can be taken multiple times. In this regard, for each maximization step, we need to construct the auxiliary set $X'$ from scratch and we do not need the previously generated auxiliary sets. However, in our few-sample training scenarios where the number of data samples is small, we can keep the auxiliary samples.

\subsection{Backward KD for NLP Applications} 
\label{nlp}
\begin{figure}[tb]
\centering
    \includegraphics[width=0.4\textwidth]{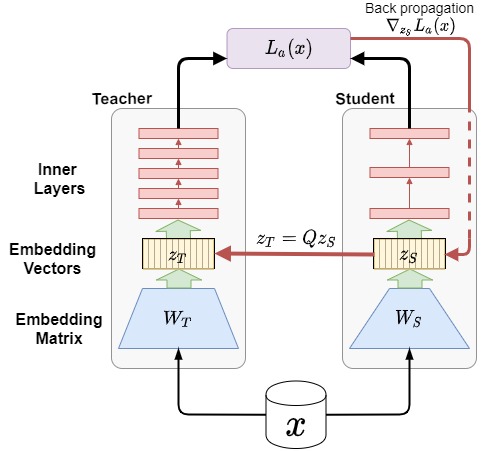}
    \caption{{General procedure of utilizing auxiliary samples in NLP models. Here $x$ is the one-hot vector of input tokens, $W$ is the embedding matrix, and $z$ is the embedding vector of $x$.}}
    \label{NLP}
\end{figure}

It is not trivial how to deploy the introduced backward KD approach (i.e. calculating $\nabla_x \mathcal{L}_{BKD}$ for discrete inputs) when data samples come from a discrete domain, such as NLP applications. Here, we propose a solution to how this technique can be adapted for the NLP domain. 
For neural NLP models, first, we pass the one-hot vectors of the input tokens to the so-called embedding layer of neural networks. Then, these one-hot vectors are converted into embedding vectors (see Fig.~\ref{NLP}). 
The key for our solution is that embedding vectors of input tokens are not discrete and we can take the gradient of loss function {w.r.t} the embedding vectors $z$. But the problem is that, in the KD algorithm, we have two networks with different embedding sizes (see Fig.~\ref{NLP}). 
To address this issue, we can take the gradient of the loss function w.r.t one of the embedding vectors (here student embedding vector $z_S$). However, then we need a transformation matrix like $Q$ to be able to derive the corresponding embedding vector $z_T$ for the teacher network form $z_S$. 
\begin{equation}
    z_T = Q z_S
 \label{IV}
\end{equation}
We can show that the transform matrix $Q$ is equal to the following equation:
\begin{equation}
    Q = W_T W_S^T (W_S W_S^T )^{-1}
    \label{V}
\end{equation}
where in this equation $W_S^T (W_S W_S^T )^{-1}$ is the pseudo inverse of $W_S$ embedding matrix. We refer you to the Appendix to see the proof of this derivation. Therefore, to obtain the auxiliary samples, we can take the gradient of the $\mathcal{L}_{BKD}$ loss function w.r.t the student embedding vector $z_S$. Then by using equations~\ref{IV} and \ref{V}, we can re-construct $z_T$ during the steps of data perturbation as following. 
\begin{equation}
 \begin{split}
&     z_S^{i+1} = z_S^i+\eta {\nabla_z}_S \mathcal{L}_{BKD} \\
 &    z_T^{i+1} = W_T W_S^T (W_S W_S^T )^{-1} z_S^{i+1}
 \end{split}
 \label{IV}
\end{equation}

\section{Experiments and Results}
We designed five experiments to evaluate our proposed method.
The first experiment is on synthetic data in order to visualize the idea behind our technique. The second and third experiments are on the image classification tasks and the last two experiments are in NLP. 
For all of these experiments, we followed the general procedure illustrated in algorithm 1 in the Appendix. For NLP experiments, we applied the method explained in section~\ref{nlp} (see algorithm 2 in the Appendix for more details).  We summarize the procedure of our experiments in the following. 

\textbf{Pre-training Step}: We train the student network based on the original KD procedure for a few epochs ($e$ epochs). In this step, the student network will get close to the teacher network around the given training samples and will diverge from the teacher in some other areas.

\textbf{Iterative Min Max Step}: We do the following two steps iteratively for several epochs ($h$ epochs) :\\
1) Using the pre-trained student network and the trained teacher network, we use the proposed maximization step in \ref{maximization} for generating an auxiliary dataset. \\
2) Combine the auxiliary data with the training dataset and train the student network based on the augmented dataset using the original KD procedure for $e$ epochs again.
 
\textbf{Fine-tuning Step}: Finally, fine-tune the student network using original KD only based on the train samples for $e$ epochs again.
The reason for this step is that, although during the previous step the student network has been got close to the teacher network in general since the student has a limited amount of parameter, it might not be able to completely converge to the teacher network using all augmented data samples. On the other hand, since the given data points are more important than the auxiliary points, then during the last step, we only train the student based on the given dataset in order to have the maximum match between student and teacher over the given data samples in the end. 

\subsection{Synthetic data experiment}
\begin{figure}[bt]
 \centering
  \subfloat[]{\label{figur:2}\includegraphics[width=0.34\textwidth]{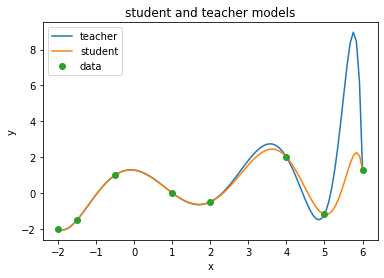}}
  \subfloat[]{\label{figur:1}\includegraphics[width=0.34\textwidth]{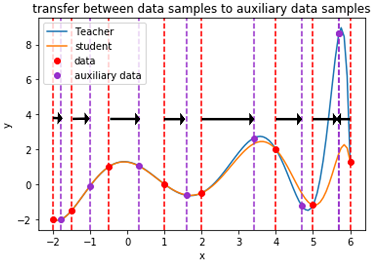}}
  \subfloat[]{\label{figur:2}\includegraphics[width=0.34\textwidth]{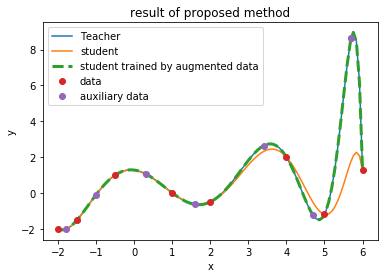}}
  \label{figur}\caption{Visualizing the generation of auxiliary samples and their utilization in training the student model.}
  \label{synthetic}
\end{figure}

For visualizing our technique and showing the intuition behind it, we designed a very simple experiment to show how the proposed method works over a synthetic setting. In this experiment, we consider a polynomial function of degree 20 as the trained teacher function.
Then, we considered 8 data points on its surface as our data samples to train a student network which is a polynomial function from degree 15 (see Fig.~\ref{synthetic}-(a)). 
As it is depicted in this figure, although the student model perfectly fits the given data points, it diverges from the teacher model in some areas between the given points. After applying our backward KD method, we can generate some auxiliary samples in the diverged areas between the teacher and student models in Fig.~\ref{synthetic}-(b). Then, we augmented the training data samples with the generated auxiliary samples and trained the student model based on this new augmented dataset. The resulting student model has achieved a much better fit on the teacher model as it is evident in Fig.~\ref{synthetic}-(c).

\subsection{MNIST classification:}
In this experiment, one of our goals was testing the performance of the proposed method in the scenario of extremely small student networks. Because of that, we considered two fully connected neural networks as student and teacher networks for the MNIST dataset classification task. The teacher network consists of only one hidden layer with 800 neurons which leads in 636010 trainable parameters. The student network was an extremely simplified version of the same network with only 5 neurons in the hidden layer. This network has only 3985 trainable parameters, which is 160x smaller than the teacher network. The student network is trained in three different ways: a) from scratch with only training data, b) based on the original KD approach with training data samples augmented by random noise, and c) based on the proposed method. As it is illustrated in table 1, the student network which is trained by using the proposed method achieves much better results in comparison with two other trained networks.

\begin{table}[ht]
\caption{Results of experiment on the MNIST dataset} 
\centering 
\begin{tabular}{c c c c} 
\hline\hline 
Model & method & \#parameters & accuracy on test set\\ [0.5ex] 
\hline 
teacher & from scratch & 636010 & 98.14 \\ 
\hline 
student & from scratch & 3985 & 87.62 \\
student & original KD & 3985 & 88.04 \\
student & \textbf{proposed method} & 3985 & \textbf{91.45} \\ [1ex] 
\hline 
\end{tabular}
\label{table:nonlin} 
\end{table}

\subsection{CIFAR-10 classification}

The second experiment is conducted on the CIFAR10 dataset with two popular network structures as the teacher and the student networks. In this experiment, we used the inception v3 ~\cite{szegedy2016rethinking} network as the teacher and mobileNet v2 ~\cite{sandler2018mobilenetv2} as the student. The teacher is approximately 9 times bigger than the student. We repeated the previous experiment on CIFAR10 by using these two networks. Table 2 shows the results of this experiment. 

\begin{table}[ht]
\caption{Results of experiment on CIFAR10 dataset} 
\centering 
\begin{tabular}{c c c c} 
\hline\hline 
Model & method & \#parameters & accuracy on test set\\ [0.5ex] 
\hline 
inception v3 (teacher) & from scratch & 21638954 & 95.41\% \\ 
\hline 
mobilenet (student) & from scratch & 2236682 & 91.17\% \\
mobilenet (student) & original KD & 2236682 & 91.74\% \\
mobilenet (student) & \textbf{proposed method} & 2236682 & \textbf{92.60\%} \\ [1ex] 
\hline 
\end{tabular}
\label{table:nonlin} 
\end{table}

\subsection{GLUE tasks}

The third experiment is designed based on General Language Understanding Evaluation (GLUE) benchmark ~\cite{wang2018glue} and roBERTa family language models~\cite{liu2019roberta, sanh2019distilbert}. The GLUE benchmark is a set of nine language understanding tasks, which are designed to evaluate the performance of natural language understanding systems. roBERTa models (roBERTa-large, roBERTa-base, and distilroBERTa) are BERT ~\cite{devlin2018bert} based language understanding pre-trained models where roBERTa-large and roBERTa-base are the cumbersome versions which are proposed in ~\cite{liu2019roberta} and have 24 and 12 transformer layers respectively. distilroBERTa is the compressed version of these models with 6 transformer layers and has been trained based on KD procedure proposed in ~\cite{sanh2019distilbert} with utilizing the roBERTa-base as the teacher. 
The general procedure in GLUE tasks is fine-tuning the pre-trained models for its down-stream tasks and the average performance score. Here, we fine-tuned the distilroBERTa model based on the proposed method by utilizing the fine-tuned roBERTa-large teacher for each of these tasks. As it is shown in table 3, the proposed method could improve the distilroBERTa performance on most of these tasks.

%

\begin{table}[ht]
\caption{Results of experiment on GLUE tasks} 
\centering 
\resizebox{\columnwidth}{!}{%
\begin{tabular}{c  c c c c c c c c c c} 
\hline\hline 
Model (Network)  & ColA & SST-2 & MRPC & STS-B & QQP & MNLI & QNLI & RTE & WNLI&  Score \\ [0.5ex] 
\hline 
roBERTa-large (Teacher)  & 60.56 & 96.33 & 89.95 & 91.75 & 91.01 &  89.11 &  93.08 &  79.06 & 56.33 & 85.82 \\
\hline 
DistilroBERTa (Student)  & 56.61 & \textbf{92.77} & 84.06 & 87.28 & 90.8 &  84.14 & \textbf{ 91.36} &  65.70 & 56.33 & 78.78  \\
\textbf{Our} DistilroBERTa (Student)  & \textbf{60.49} & 92.51 & \textbf{87.25} &  \textbf{87.56} & \textbf{91.21} & \textbf{85.1} & 91.19 & \textbf{71.11} &  56.33 & \textbf{80.30}\\
\hline 
\end{tabular}}%
\label{table:nonlin} 
\end{table}

\color{black}

\subsection{GLUE tasks with few sample points}
In this experiment, we modified the previous experiment slightly to investigate the performance of the proposed method in the few data sample scenario. Here we randomly select a small portion of samples in each data set and fine-tuned the distilroBERTa based on these samples. For CoLA, MRPC, STS-B, QNLI, RTE, and WNLI, 10\% of data samples and for SST-2, QQP, and MNLI 5\% of them in the dataset are used for fine-tuning the student model.

\begin{table}[ht]
\caption{Results of few sample experiment on GLUE tasks} 
\centering 
\resizebox{\columnwidth}{!}{%
\begin{tabular}{c  c c c c c c c c c c} 
\hline\hline 
Model (Network)  & ColA & SST-2 & MRPC & STS-B & QQP & MNLI & QNLI & RTE & WNLI&  Score \\ [0.5ex] 
\hline 
roBERTa-large (Teacher)  & 60.56 & 96.33 & 89.95 & 91.75 & 91.01 &  89.11 &  93.08 &  79.06 & 56.33 & 85.82 \\
\hline 
DistilroBERTa (Student)  & 43.82 & 91.05 & 76.96 & 81.51 & 84.92 &  75.88 & 83.94 &  52.07 & 56.33 & 71.90  \\
\textbf{Our} DistilroBERTa (Student)  & \textbf{44.11} & \textbf{91.74} & \textbf{77.20} &  \textbf{82.82} & \textbf{85.32} & \textbf{76.75} & \textbf{84.34} & \textbf{56.31} &  56.33 & \textbf{72.76}\\
\hline 
\end{tabular}}%
\label{table:nonlin} 
\end{table}

\section{Conclusion}
In this paper, we have introduced the backward KD method and showed how we can use the backward knowledge of teacher model to train the student model. Based on this method, we could easily locate the diverge areas between teacher and student model in order to acquire auxiliary samples at those areas with utilizing the gradient of the networks and use these samples in the training procedure of the student model. We showed that our proposal can be efficiently applied to the KD procedure to improve its performance. Also, we introduced an efficient way to apply backward KD on discrete domain applications such as NLP tasks.
In addition to the synthetic experiment which is performed to visualize the mechanism of our method, we tested its performance on several image and NLP tasks. Also, we examined the extremely small student and the few sample scenarios in two of these experiments. We showed that the backward KD can improve the performance of the trained student network in all of these practices. We believe that all auxiliary samples do not have the same contribution to improving the performance of the student model. Also perturbing all data samples can be computationally expensive in large datasets. 

\section*{Broader Impact}

This research provides a simple but efficient method for model compression and knowledge distillation which is easily applicable on a variety of domains in machine learning from computer vision to natural language processing with the hope of achieving better results. The proposed procedure in this work is a general procedure which can be used beside the other KD methods in order to improve their results. Since the main idea just deals with the data samples and generate more samples for better training, without any major changes in the body of other algorithms, they can use this procedure in their methods easily. It is applicable in different scenarios like extremely small student models, few data sample regimes, and zero-shot KD.

\section*{Acknowledgments}
We thank Mindspore\footnote{A new deep learning computing framework \url{https://www.mindspore.cn/}} for the partial support of this work. 
We thank the anonymous reviewers for their insightful comments.



\small
\normalem

\bibliographystyle{acm}
\bibliography{references}

\section*{Supplementary Materials}
\section{Transform matrix between student and teacher embedding}
If $W_S \in \mathbb{R}^{d_{1}\times|V|}$ be the embedding matrix of the student network and $W_T \in \mathbb{R}^{d_{2}\times|V|}$ be the embedding matrix of the teacher network, where $|V|$ is the vocabulary size and $d_1$ and $d_2$ are the embedding vector size of the student and the teacher networks respectively. If $x \in \{0,1\}^{|v|}$  be the one-hot vector of a token in a text document and if $z_S = W_{S}x$ and $z_T = W_{T}x$ be the student and teacher embedding vectors of $x$, then there exists a transform matrix $Q \in \mathbb{R}^{d_{2}\times d_{1}}$ such that:

\begin{equation}
z_T = Qz_S
\end{equation}

\paragraph{Proof:}
\begin{equation}
z_T=W_T x 
\end{equation}

\begin{equation}
z_S=W_S x  
\end{equation}

We want to find a transform matrix $Q$ such that:
\begin{equation}
W_T=Q W_S 
\label{trans}
\end{equation}
For this purpose we can solve the following optimization problem by using list square method:
\begin{equation}
\begin{split}
\underset{Q}{\text{min }}    ||W_T-Q W_S||^2 
\end{split}
\end{equation}

By solving the above optimization problem using the least squares method, we achieves the following solution for $Q$: 
\begin{equation}
\begin{split}
 Q = W_T W_s^T (W_s W_s^T )^{-1}
\end{split}
\end{equation}
Now, from Eq.~\ref{trans} we have: 	
\begin{equation}
W_T=QW_s 
\end{equation}

\begin{equation}
W_T x=Q W_s x
\end{equation}

\begin{equation}
z_T=Qz_s
\end{equation}

\section{Algorithm 1}
Algorithm 1 explains the details of the proposed method in section 3 of the paper. The input variables of our proposed KD function are the student network $S(.)$, the teacher network $T(.)$, the input dataset $X$, the number of training epochs $e$, and the number of hyper epochs $h$. In this algorithm, we assume that the teacher network $T(.)$ has trained and the student network $S(.)$ has not trained yet. Also, we assume $X'$ is the set of the augmented data samples. We first initialize $X'$ with data set X in line 3 of the algorithm. The basic idea is that each time we train the student network using the Vanilla-KD function for a few training epochs $e$ in the outer loop of line 4. Then, in line 6 first, we re-initialize $X'$ with dataset $X$ and in lines 7 to 11 we perturb data samples in $X'$ using the gradient of the loss between teacher and student iteratively in order to generate new auxiliary samples. Then in line 12 we replace $X$ with the union of $X$ and $X'$ sets. In the next iteration of the loop in line 4, Vanilla-KD function will be fed with the augmented data samples $X'$. Note that just in the first iteration, Vanilla-KD function is fed with original data set $X$.

\begin{algorithm}
    \caption{}\label{euclid}
    \begin{algorithmic}[1]
        \Function{Proposed-KD}{$S$,$T$,$X$, $e$, $h$}    
            \State \Comment{$S$ is the student network, $T$ is the teacher network, $X$ is input dataset, $e$ is \#training epochs, $h$ is \#hyper epochs}
            \State $X' \gets X$
            \For{$i = 1$ \texttt{to $h$}}
                \State \Call{Vanilla-KD}{$S$,$T$,$X'$,$e$} 
                \State $X' \gets X$
                \For{$x'$ \texttt{in $X'$}}
                    \While{converge}
                        \State $x' \gets x' + \eta \nabla_{x}||S(x') - T(x')||^2_2$
                    \EndWhile
                \EndFor
                \State $X' \gets X'\cup X$
            \EndFor
            \State \Call{Vanilla-KD}{$S$,$T$,$X$,$e$}
            \State \Return $S$
        \EndFunction
    \end{algorithmic}
\end{algorithm}

\section{Algorithm 2}
Algorithm 2, explains how to apply the proposed method in NLP tasks. This algorithm is almost similar to algorithm 1. The only main difference is in the way we feed the networks. Here instead of considering the one-hot index vectors of tokens in the text documents, we consider the embedding vectors $z_S$ and $z_T$ of the input vector $x$ (see lines 5 and 6 in the algorithm). Then we fed each of the teacher and the student networks separately using their own embedding vectors. Only in line 16 we use the transform method which is proposed in section 3.2 of the paper to transform student's perturbed embedding vectors into teacher's embedding vectors. 

\begin{algorithm}
    \caption{}\label{euclid}
    \begin{algorithmic}[1]
        \Function{Proposed-KD}{$S$,$T$,$X$, $e$, $h$}    
            \State \Comment{$S$ is the student network, $T$ is the teacher network, $X$ is input dataset, $e$ is \#training epochs, $h$ is \#hyper epochs}
            \State $W_T \gets$ \Call{Embedding-matrix}{$T$}
            \State $W_S \gets$ \Call{Embedding-matrix}{$S$}
            \State $Z_T \gets W_TX$
            \State $Z_S \gets W_SX$
            \State $Z_T' \gets Z_T$
            \State $Z_S' \gets Z_S$
            \For{$i = 1$ \texttt{to $h$}}
                \State \Call{Vanilla-KD}{$S$,$T$,$Z_T'$, $Z_S'$,$e$} 
                \State $Z_T' \gets Z_T$
                \State $Z_S' \gets Z_S$
                \For{($z_S'$, $z_T'$) \texttt{in ($Z_S'$, $Z_T'$)}}
                    \While{converge}
                        \State $z_S' \gets z_S' + \eta \nabla_{z_S}||S(z_S') - T(z_S')||^2_2$
                        \State $z_T' \gets W_T W_S (W_SW_S^T)^{-1}z_S'$
                    \EndWhile
                \EndFor
                \State $Z_S' \gets Z_S'\cup Z_S$
                \State $Z_T' \gets Z_T'\cup Z_T$
            \EndFor
            \State \Call{Vanilla-KD}{$S$,$T$,$Z_T$, $Z_S$,$e$}
            \State \Return $S$
        \EndFunction
    \end{algorithmic}
\end{algorithm}

\end{document}